\documentclass[conference]{IEEEtran}
\IEEEoverridecommandlockouts
\usepackage{cite}
\usepackage{amsmath,amssymb,amsfonts}
\usepackage{graphicx}
\usepackage{textcomp}
\usepackage{xcolor}
\def\BibTeX{{\rm B\kern-.05em{\sc i\kern-.025em b}\kern-.08em
    T\kern-.1667em\lower.7ex\hbox{E}\kern-.125emX}}
\usepackage{hyperref}
\usepackage{booktabs}
\usepackage{subfig}
\usepackage{algpseudocode}
\usepackage{multirow}
\usepackage{pifont}

\newcommand\mydots{\hbox to 1em{.\hss.\hss.}}

\begin{document}

\title{Independent Modular Networks\\
\thanks{$^\dag$ School of Computer Science, University of Adelaide, Australia\\
$^*$ College of Engineering and Computing, University of South Carolina, Columbia, South Carolina, USA.}
}
\author{\IEEEauthorblockN{Hamed Damirchi$^\dag$}
\IEEEauthorblockA{hamed.damirchi@adelaide.edu.au}
\and
\IEEEauthorblockN{Forest Agostinelli$^*$}
\IEEEauthorblockA{foresta@cse.sc.edu}
\and
\IEEEauthorblockN{Pooyan Jamshidi$^*$}
\IEEEauthorblockA{pjamshid@cse.sc.edu}

}

\maketitle

\begin{abstract}
Monolithic neural networks that make use of a single set of weights to learn useful representations for downstream tasks explicitly dismiss the compositional nature of data generation processes. This characteristic exists in data where every instance can be regarded as the combination of an identity concept, such as the shape of an object, combined with modifying concepts, such as orientation, color, and size. The dismissal of compositionality is especially detrimental in robotics, where state estimation relies heavily on the compositional nature of physical mechanisms (e.g., rotations and transformations) to model interactions. To accommodate this data characteristic, modular networks have been proposed. However, a lack of structure in each module's role, and modular network-specific issues such as module collapse have restricted their usability. We propose a modular network architecture that accommodates the mentioned decompositional concept by proposing a unique structure that splits the modules into predetermined roles. Additionally, we provide regularizations that improve the resiliency of the modular network to the problem of module collapse while improving the decomposition accuracy of the model.

\end{abstract}


\section{Introduction}
\label{sec:intro}
Representation learning using monolithic models (models that use a single set of weights for every input) typically involves passing all data samples through the model and extracting features after training on pretext tasks deemed helpful for downstream tasks. However, this approach ignores the compositional characteristics of data generation processes. As an example of this common characteristic for images of objects, every object can be described as a composition of the object's shape, color, size, orientation, texture, etc. Instead of learning these concepts separately while considering the compositional nature of each possible combination of the mentioned concepts, monolithic models attempt to directly extract high-level features without any constraints that would impose such a structure explicitly.


Modular neural networks \cite{Kirsch2018ModularNL} was proposed as a potential solution that considers this compositional characteristic, where a set of modules instead of a singular neural network is used to extract features from any given input. Generally, the feature extraction process for these models proceeds using a scoring method to determine which module or set of modules from the available modules should process the input. Then, the input is passed through the chosen modules. In the case where multiple modules were chosen, a combination method, such as adding features, is chosen to combine the features extracted by each module. 

Regardless of the module selection approach, while modular networks take a step towards solutions considering the mentioned compositional characteristics of data, one particular structural element is still missing from the currently available works. To have a compositional structure, there is a need for an identity state to be altered through combination with compositional concepts. We define this identity state as one of the true generative factors of the dataset that does not require another factor to be defined. Using the previous example of images of objects, one can consider the shape of the object (a physical concept) as the identity state that can be modified by compositional concepts (non-physical concepts) such as rotations, color, and scaling. In this example, while the object's shape can be defined separately from the other factors, the remaining factors require a shape to be defined in a hierarchical manner beforehand. Therefore, we propose structural changes to modular networks that allow the learning of two separate sets of modules, where one set of modules automatically learns a notion of identity, and the other set only encodes the compositional concepts. In particular, this is done by only allowing the compositional modules to observe the input while the identity modules do not. Instead, the identity modules will use a set of learnable parameters to output a static state that is modified using compositional modules. The reason for this change is that the identity modules will now have to learn a specific, unchangeable concept that remains static among large portions of the dataset.

A problem that modular networks commonly deal with is when the model routes all the inputs through the same module or set of modules. Different works propose varying solutions, such as regularizing the router to choose diverse modules or modifying the way modules are combined during feature extraction. In this work, we propose to solve this problem indirectly by imposing independence constraints for the features extracted by each module and among the compositional and identity modules. By penalizing correlations between the concepts embedded by each module, this constraint would force modules to learn to embed information not considered by other modules, which leads to the decomposition of the concepts present in the dataset into separate modules.


\begin{itemize}
    \item Propose a new architecture for modular networks that promotes learning features that are in line with the compositional nature of data generation processes 
    \item Propose an independence-based solution that indirectly solves module collapse.
    \item Provide experiments showcasing the automatic decomposition capabilities of the proposed approach in the extraction of the identity states.
\end{itemize}

\section{Related Work}
\label{sec:relwork}
Algorithmically, the previous works on modular networks share a few structural traits, such as the usage of a router and the recursiveness of the application of modules to the input. Meanwhile, other details such as the training method, learnability of different components of the methods, and the application differ between the works in the literature. In the following, a summary of the literature alongside the distinction between each work and ours is delineated.

In \cite{Kirsch2018ModularNL}, a router and a set of modules are trained using a generalized expectation-maximization (EM) algorithm. For each input, the router selects a set of modules from a library of modules to process the input. The outputs from selected modules are then combined (either through concatenation or addition) to compute the output of the layer. While a modular approach, this method does not impose any structural constraints on the modules to decompose features into identity and compositional states.

While making use of reinforcement learning algorithms instead of EM, \cite{chang2018automatically} allows the routing module to decide when to stop processing the original input rather than using a predetermined number of recursions. While the decomposition of the input into multiple atomic representations using multiple modules is similar to our work, we do not constrain our modules to encode specific information such as rotation, and the learning is done automatically during the training.

Each of the mentioned works deals with a problem called module collapse, where during training, the routing algorithm routes all the inputs through only a few of the modules. \cite{Kirsch2018ModularNL} uses a different number of M-step compared to the E-step in the EM algorithm to mitigate this issue. On the other hand, \cite{rosenbaum2018routing} finds that using multiple agents per task mitigates the module collapse problem. Even though previous works have, in some cases, shown that module collapse is prevented using various regularization techniques, without an independency constraint, diversity in usage of modules may not mean that module collapse does not occur since multiple modules might still encode the same concept. Conventional module collapse detection methods cannot detect this. In contrast, our proposed independence-based approach solves this problem indirectly by preventing the same concept from being encoded by multiple modules.


\section{Proposed approach}
\label{sec:approach}
\begin{figure}
\centering
\includegraphics[width=0.5\linewidth]{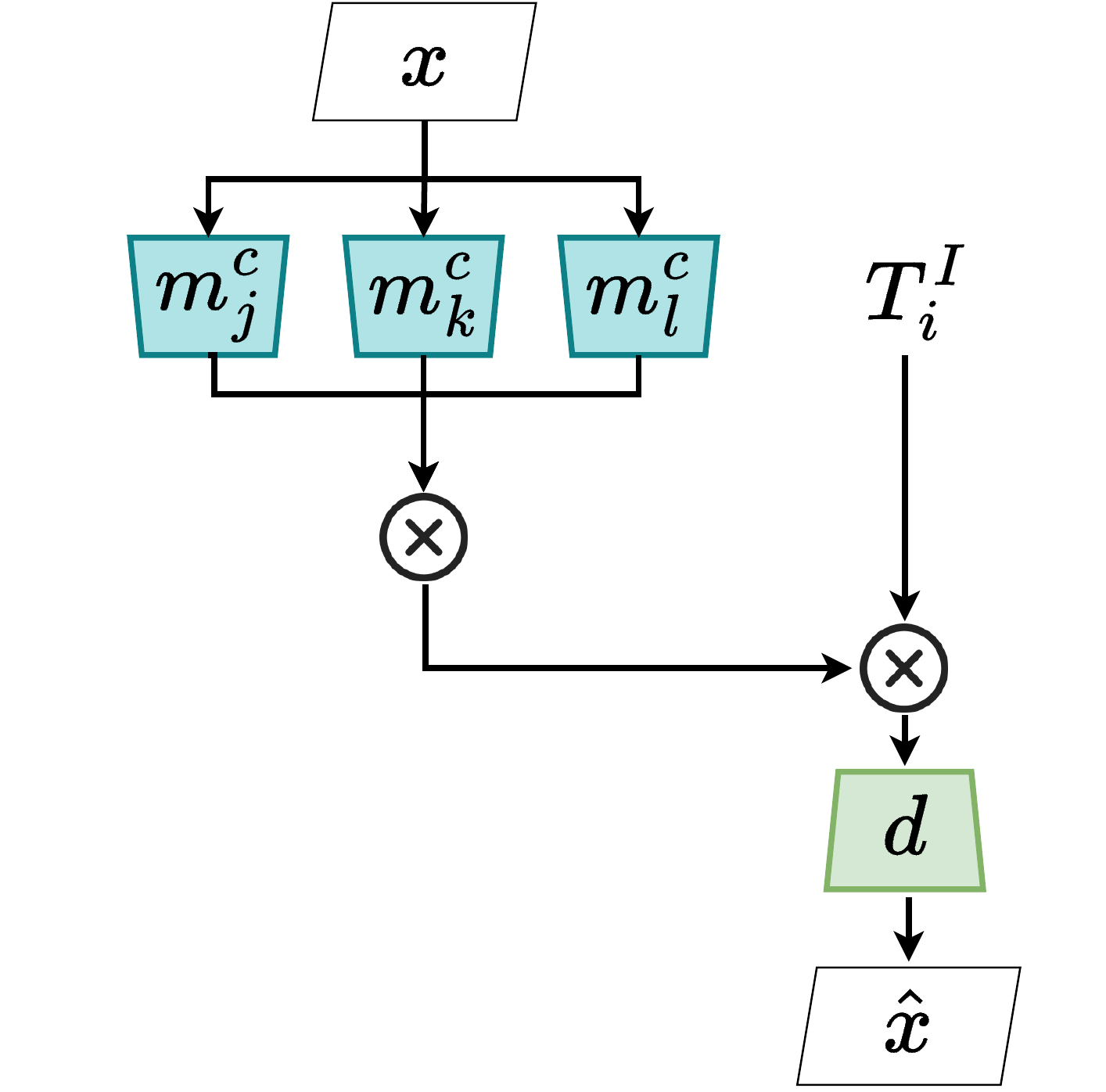}
\caption{The general structure of the proposed approach}
\label{fig:general}
\vspace{-0.5cm}
\end{figure}

The general architecture of the proposed approach is shown in Fig. \ref{fig:general}. The pool of available modules can be split into two categories, where one group of modules is only used for learning compositional concepts, and modules from this group are able to observe the input. The other modules are not able to observe the input and output of a learnable set of parameters in the shape of a transformation matrix to be combined with compositional modules. Each compositional module's architecture is similar to that of the encoder of Lie group Variational Autoencoders (LVAE) \cite{xinqi2021commutative}. Modules designated $T^I$ represent the learnable identity matrices we name identity modules, while $m^c$ designates compositional modules. While the proposed approach can be used with any set of data with compositional characteristics, we use an image reconstruction application in this paper. Therefore, a decoder is required to reconstruct the image using the features extracted by the modules. This module is designated by $d$ in Fig. \ref{fig:general}. To train the model, we first output a reconstruction of the input image using every combination of the available modules. Then, a winning combination is chosen by choosing the combination that provides an output with the lowest loss. In this section, we give a brief background of LVAEs. Then, feature extraction and the reconstruction processes are delineated. Finally, the regularizations and loss calculation methods are provided.

\subsection{Lie-group VAE}
\label{sec:subsectionlvae}
LVAE \cite{xinqi2021commutative} is a variational autoencoder that extracts a vector of disentangled representations for the input image, while the same vector is an element of a Lie group represented using the tangent space of the group. This way, smooth equivariant representations are learned where the encoder predicts values along the axes of the Lie algebra vector that are converted to their matrix representations through exponential mapping. In feature extraction, the input is first passed through an encoder typically chosen as a CNN for feature extraction from images and a distribution is inferred over the position of the latent variables. After sampling from this distribution, the latent variables are used alongside the learnable Lie algebra bases to infer the element of the group representing the input through the exponential operator. This process is formulated as follows \cite{xinqi2021commutative}.
\begin{equation}
    \begin{split}
        \mu, \sigma &= g(x),\; z = \mu+\sigma\epsilon, \\
        T &= exp(zA),\; \hat{x} = d(T) \\
    \end{split}
\end{equation}
Where $g$ and $d$ are the encoders and the decoder of this network, $z$ is the latent vector, and $A$ is the learnable Lie algebra bases. Additionally, $T\in R^{u\times x}$ represents the extracted features as a transformation matrix, i.e., an element of the group representing the input. To recover disentangled representations, regularizations are applied to the network. \cite{xinqi2021commutative} showed that by imposing $A_{i} A_{j}=0, \forall i \neq j$, the following would be satisfied.
\begin{equation}
H_{i j}=\frac{\partial^{2} g(t)}{\partial z_{i} \partial z_{j}}=0
\label{eq:hessiandisen}
\end{equation}
where $A_i$ is the $i^{th}$ Lie algebra basis and $H$ represents the Hessian matrix. If (\ref{eq:hessiandisen}) is satisfied for every $i$ and $j$ when $i\neq j$, then the changes in the output of the network with respect to a latent variable are not dependent on the changes of other latent variables. This ensures that independent concepts are encoded into each latent variable during training.

\subsection{Modular feature extraction}
\label{sec:featureextraction}
To extract the features from an image, every compositional module first processes the input image as follows:
\begin{equation}
        T^c_i = m^c_i(x)
\end{equation}
where $T^c_i$ represents the transformation features extracted by the $i^{th}$ compositional module. Thereafter, every combination of the compositional features is calculated, denoted as:
\begin{equation}
    \mathbf{T}^c = \{T^c_1, \mydots, T^c_m, T^c_1T^c_2, \mydots, T^c_{m-1}T^c_m, \mydots , T^c_1 \mydots T^c_{m-1}T^c_m\}
\end{equation}
where $\mathbf{T}^c$ represents the set of all combinations of the compositional features, and $m$ is the number of compositional modules set manually at the start of training. Now, to obtain the combined representation of the compositional features with the identity transformations, every element of the set $\mathbf{T}^c$ is multiplied by the learnable transformation matrix representing each identity module as follows:
\begin{equation}
    \mathbf{T} = \{T_{c_i}T_{I_j} \:|\; T_{c_i} \in \mathbf{T}^c \;,\; T_{I_j} \in \mathbf{T}^I\},
    \label{eq:combination}
\end{equation}
where $\mathbf{T}^I=\{T^I_1,\cdots, T^I_n\}$ represents the set of learnable transformations from every identity module where the pool of available modules consists of $n$ identity modules. Every element of the set $\mathbf{T}$ will be considered the candidate representation for input $x$ until the combination that produces the best output is chosen based on the scoring criteria depicted in Section \ref{sec:scoring}. But first, we need to produce a reconstruction based on every combined feature in $\mathbf{T}$ so that every element of this set can be evaluated in the image space. In this work, we use a single decoder network for every feature to get a set of reconstructions as follows:
\begin{equation}
    \hat{\mathbf{X}} = \{d(T_i) \;|\; T_i\in \mathbf{T}\},
\end{equation}
where $d$ represents the decoder network.
\subsection{Scoring, regularization, and loss function}
In order to update the weights of the modules, two steps remain. The winning combination needs to be chosen first. Then, based on the output from the winning combination, the loss will be calculated, and after the computation of the gradients, the weights of the modules responsible for the generation of the winning output will be updated.
\subsubsection{Scoring the output from each combination}
\label{sec:scoring}
Three criteria are used to score the reconstruction and the features extracted using each combination of modules. First, the reconstruction is evaluated using image mean-squared-error loss
\begin{equation}
    \mathcal{L}^{img}_{i} = \|x-\hat{x}_{i}\|,
    \label{eq:imgmse}
\end{equation}
where $\hat{x}_i$ is the $i^{th}$ element of $\hat{\mathbf{X}}$. The second criterion quantifies the level of independence between the features extracted by each module, alongside the independency of the features between the different modules. By promoting intra-module independency, the modules that output higher-quality latent variables are prioritized. This is while inter-module independency constraints incentivize combinations of modules that embed the least amount of correlated information with respect to other modules in the combination. To quantify intra-module independence (which will also be used in the loss function later), we use the same approach as \cite{xinqi2021commutative} where the changes of the gradient of the output with respect to a single element of the latent vector are computed with respect to every other element of the same vector. We extend this formulation to inter-module independence as well. The following shows the formulation for both independence quantification methods.
\begin{equation}
    \mathcal{L}^{ind} = \sum^{}_{i,j}\frac{\partial^2\hat{x}}{z_iz_j}+\sum^{}_{i,j}\frac{\partial^2\hat{x}}{z^\prime_iz^\prime_j}
    \label{eq:indep}
\end{equation}
where $z_i,z_j\in Z_k^I$ and $z^\prime_i\in Z_k^I, z^\prime_j\in Z_k^I{^\prime}$. Additionally, $Z_k^I$ represents the latent variables extracted by the $k^{th}$ module in the combination under evaluation. Therefore, the first term on the right-hand side of (\ref{eq:indep}) quantifies how correlated the variables in the latent vector of one module are while the second term quantifies how correlated the variables of one module are compared to that of another module in the combination under evaluation. The final term of the scoring function is the Kullback-Leibler Divergence (KLD) between the predicted and the true posterior distribution over the latent used as a part of the originally proposed VAE \cite{originalvae2013kingma} loss function. Putting this scoring formula together, we have the following equation.
\begin{equation}
    S_i = -(\mathcal{L}^{img}_i+\mathcal{L}^{ind}_i+\sum_j\mathcal{L}^{KL}_{ij})
    \label{eq:scoring}
\end{equation}
where $S_i$ determines the score for the $i^{th}$ combination from (\ref{eq:combination}) and $j$ represents the KL divergence for the latent variable of the $j^{th}$ module in combination $i$. To choose the winning combination, the one with the maximum score is selected and the loss is calculated for this winning combination.
\subsubsection{Calculating the loss for the winning combination}
\label{sec:losscalc}
With the winning combination selected, the only step left is to calculate the loss and update the weights of the modules and the learnable parameters representing the learned identity of the winning combination. To this end, the following is used.
\begin{equation}
    \mathcal{L}_i = \mathcal{L}^{img}_i+\mathcal{L}^{ind}_i+\sum_j\mathcal{L}^{KL}_{ij}+\mathcal{L}^{extra}_i
    \label{eq:fullloss}
\end{equation}
where the same image, independence, and KLD loss are re-used by adding an extra LVAE-specific loss term that is not included in the scoring process and is only used to ensure training stability in \cite{xinqi2021commutative}. So far, the loss function does not include any terms that would give the modules information about what the encoded identity matrices should represent. To explore the effects of the introduction of such an element, we use an additional loss and compare the results to the case where (\ref{eq:fullloss}) is used. This term aims to guide the modules toward learning a specific ground truth factor as identity representations. To do this, we introduce a classifier with a fully-connected architecture trained to predict the shape of the input based on the identity transformation of the winning combination. Moreover, to ensure the imposed identity remains unchanged by the compositional modules, we add another loss that lowers the $L^2$ distance between the classifier features of the identity and the combined representations that we name ID classifier loss.
\begin{figure}%
\centering
\subfloat[Example of the 3dshapes dataset]{%
\label{fig:examples3dshapes}%
\includegraphics[width=0.84\linewidth]{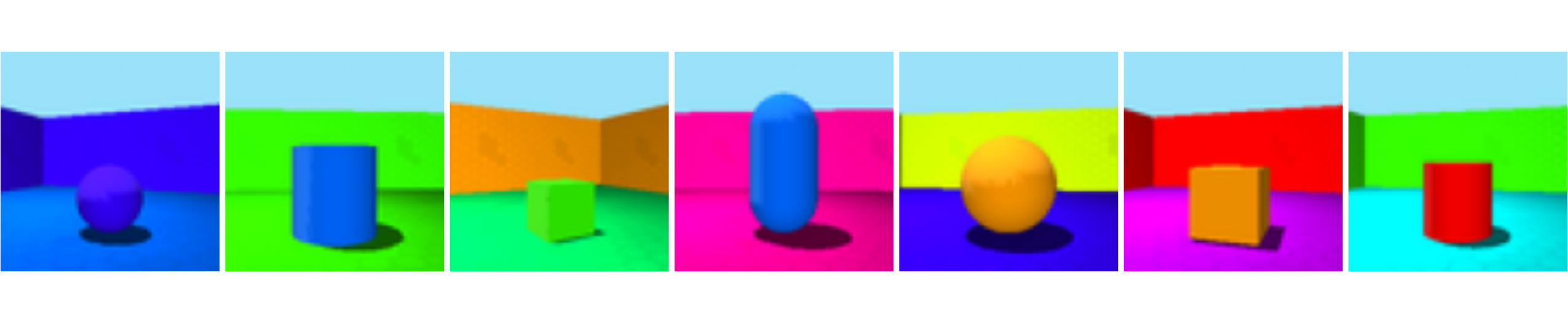}}%
\vfill\vspace{-0.4cm}
\subfloat[Learned identity reconstruction on 3Dshapes]{%
\label{fig:ident_noidcls}%
\includegraphics[width=0.85\linewidth]{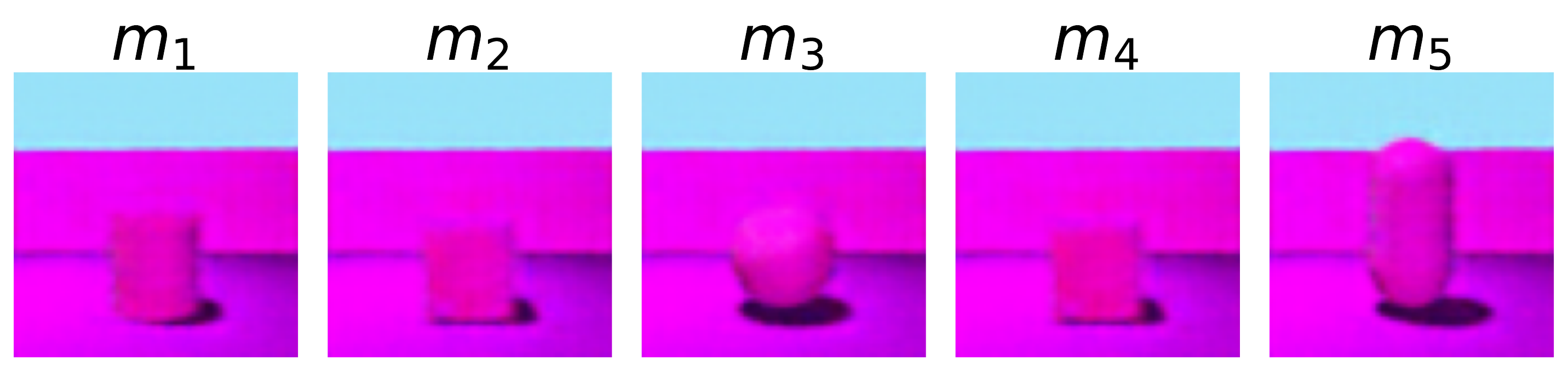}}%
\vfill\vspace{-0.4cm}
\subfloat[Learned identity reconstruction on 3Dshapes with ID classifier]{%
\label{fig:ident_idcls}%
\includegraphics[width=0.85\linewidth]{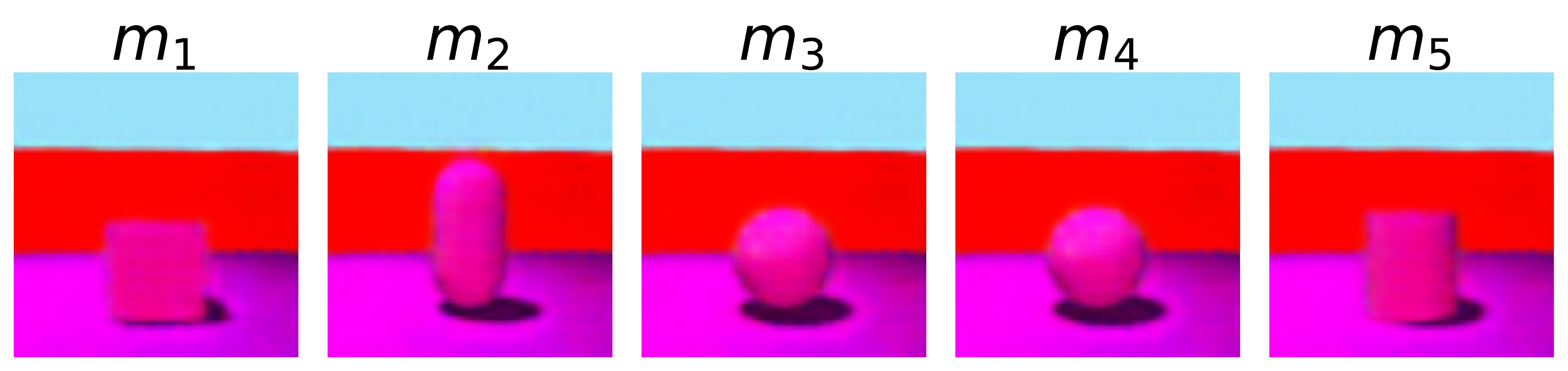}}%
\caption{Reconstructions of identity representations}
\label{fig:ident_recons}
\vspace{-0.5cm}
\end{figure}
\section{Experiments}
\subsection{Datasets}
We used the 3Dshapes \cite{3dshapes18} dataset, a synthetically generated dataset of 3D objects in the middle of a room. The variable factors used to generate this dataset consist of the camera's orientation with respect to the object, shape, and scale, and the color of the walls, object, and floor. The objects in this dataset are capsules, spheres, cylinders, and cubes. A few example images of this dataset are shown in Fig. \ref{fig:examples3dshapes}. A pool of 5 compositional modules and 5 learnable identity representations is available for every training session.
\subsection{What are the learned identities?}
To visualize what the identity representations learn, we use the decoder without any input images to reconstruct the image represented by each learned identity. Fig. \ref{fig:ident_noidcls} shows one image per identity reconstruction. Note that the loss used for this training session is based on (\ref{eq:fullloss}), and the ID classifier mentioned in Section \ref{sec:losscalc} is not used. Based on this figure, the 4 shapes available in the dataset are split among the 5 modules, where one of the shapes (cube) is learned by 2 modules. Therefore, it is evident that the proposed identity-preserving approach to modular networks is able to decompose the shapes into different learnable representations automatically without the need for labels for the different shapes. Additionally, despite the presence of 6 different ground truth generative factors, the learned representations are the shape of the object and not any of the other 5 factors. Fig. \ref{fig:ident_idcls} visualizes the reconstructions when an ID classifier is used. Similar to the previous reconstructions, one shape is learned twice, and the decomposition is based on shapes.
\begin{table}[t]
\centering
\resizebox{0.9\linewidth}{!}{
    \setlength{\tabcolsep}{3pt}
    \begin{tabular}{lllllll} \toprule
        {ID Cls.} & {GT Shape} & {$m_1 (\%)$} & {$m_2 (\%)$} & {$m_3 (\%)$} & {$m_4 (\%)$} & {$m_5 (\%)$} \\ \midrule
        \multirow{4}{*}{\ding{55}} & Cube & 28.03 & 28.80 & 0 & 43.17 & 0 \\
        & Cylinder & 100.0 & 0 & 0 & 0 & 0 \\
        & Sphere & 0 & 0 & 100.0 & 0 & 0 \\
        & Capsule & 40.32 & 0 & 0 & 0 & 59.68 \\ \midrule
        \multirow{4}{*}{\ding{51}} & Cube & 93.62 & 0 & 0 & 0 & 6.377 \\
        & Cylinder & 0 & 0 & 0 & 0 & 100.0 \\
        & Sphere & 0 & 0 & 12.33 & 87.66 & 0 \\
        & Capsule & 0 & 77.80 & 13.34 & 7.992 & 0 \\ \bottomrule
        
    \end{tabular}
}
\caption{Shape decomposition evaluation}
\label{tab:shape_decomp}
\vspace{-0.5cm}
\end{table}
\subsection{What modules are responsible for what shapes?}
The previous experiment confirms that the learned identities represent shapes when reconstructed. However, the same experiment needs to be performed when inputs are present to study the effects of compositional modules on these identities. To this end, we separate the images in the test split with respect to their ground truth shapes. These images are then passed through the dataset, and the winning combination for each image is stored. Using this data, Table \ref{tab:shape_decomp} is created where each entry in the table quantifies the fraction of images of that row's ground truth shape that had that column's learned identity in the winning combination for that image. This table is split row-wise for the two experiments based on the usage of an ID classifier. Without an ID classifier, all images containing a cylinder are passed through $m_1$, and all images of spheres are passed through $m_3$, which shows perfect decomposition for the two shapes. However, due to a lack of regularization on the compositional modules, these  modules modify the shape of the learned identity alongside other variable factors, causing an imperfect decomposition for shapes cube and capsule. With the introduction of the ID classifier, the decomposition is improved significantly, where a large majority of the images for each shape are passed through one module only, leaving $m_3$ with nearly no images.
\section{Conclusion}
In this work, we introduced a modular network-based model that is able to decompose the generative factors of a synthetically generated dataset into multiple compositional modules and static transformation matrices representing a physical factor present in the ground truth factors of the dataset. This is done by modifying the structure of the model and preventing a set of modules from observing the input while imposing independence constraints for each module separately and between different modules. Each identity is embedded in a single transformation matrix modifiable by the learned compositional modules. 
\bibliography{conference_101719.bib}
\bibliographystyle{IEEEtran}

\end{document}